\documentclass[10pt,a4paper]{article}

\usepackage[
  top=0.75in,
  bottom=0.85in,
  left=0.62in,
  right=0.62in
]{geometry}

\usepackage[utf8]{inputenc}
\usepackage[T1]{fontenc}
\usepackage{CJKutf8}

\usepackage{XCharter}
\usepackage[scaled=1.1]{zlmtt}
\usepackage{microtype}
\usepackage{helvet}

\usepackage{amsmath,amssymb,amsfonts}
\usepackage{mathtools}
\usepackage{bm}

\usepackage{graphicx}
\usepackage{booktabs}
\usepackage{multirow}
\usepackage{subcaption}
\usepackage{wrapfig}
\usepackage{float}
\usepackage{ulem}

\usepackage{multicol}
\usepackage{algorithm}
\usepackage{algorithmic}

\usepackage{setspace}
\usepackage{enumitem}
\setlist{nosep,leftmargin=*,topsep=2pt,parsep=1pt,itemsep=1pt}

\usepackage[table]{xcolor}

\definecolor{jluDeepBlue}{RGB}{0,51,153}
\definecolor{jluRed}{RGB}{198,40,40}
\definecolor{jluTeal}{RGB}{0,137,123}
\definecolor{jluAmber}{RGB}{255,160,0}
\definecolor{jluPurple}{RGB}{106,27,154}
\definecolor{jluLightBg}{RGB}{245,247,252}
\definecolor{jluGray}{RGB}{100,100,100}

\newcommand{\MonthYear}{\the\year-\ifnum\month<10 0\fi\the\month}

\usepackage[colorlinks=true,linkcolor=jluDeepBlue,citecolor=jluTeal,urlcolor=jluRed]{hyperref}
\usepackage{cleveref}
\usepackage[numbers,sort&compress]{natbib}
\usepackage{fancyhdr}
\usepackage{tikz}
\usetikzlibrary{calc}
\usepackage{tcolorbox}
\tcbuselibrary{skins,breakable}
\usepackage{titlesec}
\usepackage{caption}
\titleformat{\section}
  {\normalsize\rmfamily\bfseries\color{jluDeepBlue}}
  {\colorbox{jluDeepBlue}{\textcolor{white}{\;\thesection\;}}}{0.5em}{}
  [\vspace{1pt}{\color{jluDeepBlue}\titlerule[0.8pt]}]

\titleformat{\subsection}
  {\normalsize\rmfamily\bfseries\color{black}}
  {\thesubsection}{0.5em}{}

\titleformat{\subsubsection}
  {\small\sffamily\bfseries\color{black}}
  {\thesubsubsection}{0.5em}{}

\titlespacing*{\section}{0pt}{1.0em}{0.4em}
\titlespacing*{\subsection}{0pt}{0.7em}{0.25em}
\titlespacing*{\subsubsection}{0pt}{0.5em}{0.15em}

\newcommand{\logobadge}[2]{%
  \tikz[baseline=-0.5ex]{%
    \node[fill=#1, text=white, rounded corners=3pt,
          inner xsep=6pt, inner ysep=3pt,
          font=\small\sffamily\bfseries]{#2};}}

\newcommand{\logoimg}[4][1.9em]{% [height]{stem}{color}{fallback-text}
  \IfFileExists{logo/#2.pdf}%
    {\includegraphics[height=#1]{logo/#2}}%
    {\IfFileExists{logo/#2.png}%
      {\includegraphics[height=#1]{logo/#2}}%
      {\IfFileExists{logos/#2.pdf}%
        {\includegraphics[height=#1]{logos/#2}}%
        {\IfFileExists{logos/#2.png}%
          {\includegraphics[height=#1]{logos/#2}}%
          {\IfFileExists{#2.pdf}%
            {\includegraphics[height=#1]{#2}}%
            {\IfFileExists{#2.png}%
              {\includegraphics[height=#1]{#2}}%
              {\logobadge{#3}{#4}}}}}}}}

\newtcolorbox{highlightbox}[1][]{%
  enhanced, breakable,
  colback=jluLightBg,
  colframe=jluDeepBlue,
  coltitle=white,
  fonttitle=\small\sffamily\bfseries,
  boxrule=0.6pt,
  arc=3pt,
  left=5pt, right=5pt, top=4pt, bottom=4pt,
  title=#1}

\begin{document}
\thispagestyle{fancy}

% ======================  TITLE BLOCK (full width) ============
% --- title ---
\begin{center}
  {\fontsize{18}{24}\selectfont\rmfamily\bfseries\color{black}%
    OmniStyle2: Learning to Stylize by Learning to Destylize\par}
  \vspace{1pt}
  
  \vspace{6pt}

  % --- authors ---
  {\normalsize\rmfamily
    \textbf{Ye Wang}\textsuperscript{1}\quad
    \textbf{Zili Yi}\textsuperscript{2}\quad
    \textbf{Yibo Zhang}\textsuperscript{1,3}\quad
    \textbf{Peng Zheng}\textsuperscript{1,3}\quad
    \textbf{Xuping Xie}\textsuperscript{1}\quad
    \textbf{Jiang Lin}\textsuperscript{2}\\[2pt]
    \textbf{Yijun Li}\textsuperscript{4}\quad
    \textbf{Yilin Wang}\textsuperscript{4,$\star\dagger$}\quad
    \textbf{Rui Ma}\textsuperscript{1,5,$\star$}}\\[4pt]
  % --- affiliations ---
  {\footnotesize\rmfamily\color{black}
    \textsuperscript{1}Jilin University\quad
    \textsuperscript{2}Nanjing University\quad
    \textsuperscript{3}Shanghai Innovation Institute\\[1pt]
    \textsuperscript{4}Adobe\quad
    \textsuperscript{5}Engineering Research Center of Knowledge-Driven Human-Machine Intelligence, MOE, China}\\[2pt]
  {\footnotesize\rmfamily\color{black}
    $\star$ Corresponding authors \quad\quad $\dagger$ Project lead}
\end{center}

\vspace{-1em}
\noindent{\color{jluDeepBlue}\rule{\textwidth}{0.7pt}}
\vspace{-1em}

\begin{figure}[H]
    \centering
    \includegraphics[width=\linewidth]{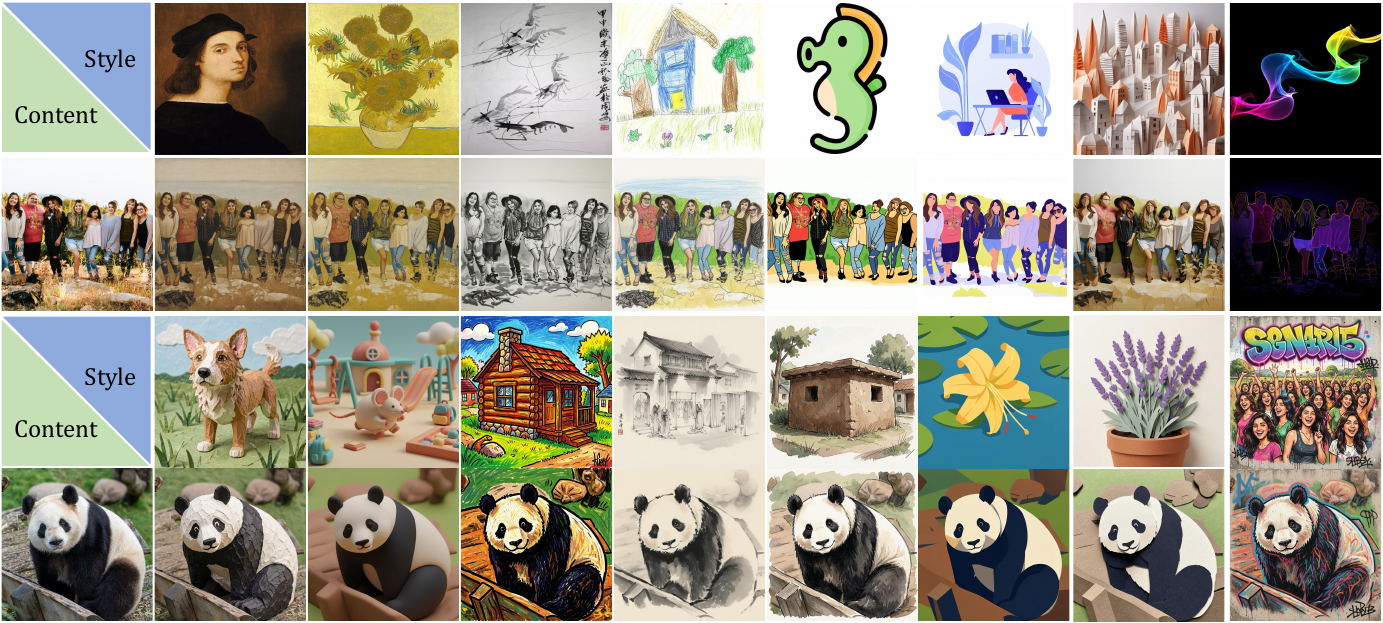}
    % \vspace{-2em}
    \caption{
      \textbf{Diverse stylization results of our method.} Our framework can generates high-fidelity stylization results across a wide range of artistic styles at 1K resolution.
    }
    % \vspace{-1em}
    \label{fig:vs4o}
\end{figure}

% --- abstract (full width) ---
\begin{center}
\begin{tcolorbox}[
  enhanced,
  width=0.92\textwidth,
  colback=jluDeepBlue!6!white,
  colframe=jluDeepBlue!35!white,
  boxrule=0.6pt,
  arc=3pt,
  left=6pt,right=6pt,top=5pt,bottom=5pt
]
  {\small
  \noindent{\sffamily\bfseries\color{jluDeepBlue}Abstract\quad}%
  This paper introduces a scalable paradigm for supervised style transfer by inverting the problem: instead of learning to stylize directly, we learn to destylize, reducing stylistic elements from artistic images to recover their natural counterparts and thereby producing authentic, pixel-aligned training pairs at scale. To realize this paradigm, we propose DeStylePipe, a progressive, multi-stage destylization framework that begins with global general destylization, advances to category-wise instruction adaptation, and ultimately deploys specialized model adaptation for complex styles that prompt engineering alone cannot handle. Tightly integrated into this pipeline, DestyleCoT-Filter employs Chain-of-Thought reasoning to assess content preservation and style removal at each stage, routing challenging samples forward while discarding persistently low-quality pairs. Built on this framework, we construct DeStyle-350K, a large-scale dataset aligning diverse artistic styles with their underlying content. We further introduce BCS-Bench, a benchmark featuring balanced content generality and style diversity for systematic evaluation. Extensive experiments demonstrate that models trained on DeStyle-350K achieve superior stylization quality, validating destylization as a reliable and scalable supervision paradigm for style transfer. Our project page: \href{https://wangyephd.github.io/projects/DeStyle/index.html}{https://wangyephd.github.io/projects/DeStyle/index.html}}
\end{tcolorbox}
\end{center}

\vspace{0.4em}

\section{Introduction}
\label{sec:introduction}

Style transfer \cite{gatys2016image}, which aims to render an image in a specific artistic style while preserving its underlying semantic content, has witnessed remarkable progress, evolving from early optimization methods \cite{gatys2016image,gatys2017controlling,kolkin2019style} to highly expressive diffusion-based solutions \cite{wang2024instantstyle,wang2024instantstyleplus,xing2024csgo,gao2024styleshot,sohn2023styledrop}. However, it remains fundamentally ill-posed due to the absence of definitive ``ground-truth'' stylization for a given content-style pair. Most prior works attempt to circumvent this from a model-centric perspective \cite{gatys2015neural, gatys2016image, zhang2019multimodal, kolkin2019style, gatys2017controlling, sohn2023styledrop, frenkel2024implicit, ouyang2025k, shah2023ZipLoRA, wang2025sigstyle,Chung_2024_CVPR, zhang2023prospect, voynov2023p+}. Yet, without explicit supervision, they often suffer from inaccurate style representation and uncontrollable optimization. OmniStyle \cite{wang2025omnistyle} recently pioneered a data-centric approach by synthesizing a large-scale paired dataset. However, because its targets are generated by existing style transfer models, it inevitably provides pseudo-supervision, yielding unauthentic approximations that fail to guarantee consistent stylization.

In this paper, we propose a fundamentally different path toward supervised style transfer: \textbf{learning to stylize by learning to destylize}. Instead of synthesizing stylized images from scratch, we reverse the process by automatically reducing stylistic elements to extract structure-aligned natural content from diverse artistic inputs. This paradigm establishes a reverse formulation: original, unaltered artistic images directly serve as authentic ground-truth targets, while the destylized images act solely as content inputs. By doing so, the supervision quality is strictly anchored to high-quality original art. Furthermore, any minor imperfections in the destylized content naturally serve as robust data augmentation, improving model generalization without compromising the target supervision. To scale this paradigm, we propose \textbf{DeStylePipe}, a progressive multi-stage framework that robustly extracts structure-aligned natural content through three stages: global general destylization, category-wise instruction adaptation, and specialized model adaptation. To ensure stringent data quality, we integrate \textbf{DestyleCoT-Filter}, an interpretable Chain-of-Thought evaluation mechanism. Crucially, by operating on destylized natural images rather than complex stylized outputs, this filter fundamentally aligns with the primary training distribution of MLLMs, ensuring robust and reliable quality control. Built on this framework, we construct \textbf{DeStyle-350K}, a large-scale dataset comprising 350K high-quality triplets $\langle$ \textcolor{green}{\textbf{de-stylized image}}, \textcolor{green}{\textbf{reference image}}, \textcolor{blue}{\textbf{style image}} $\rangle$\footnote{green: input; blue: ``ground-truth''} across over 500 styles and 100 semantic classes. Finally, we present \textbf{BCS-Bench}, a benchmark with balanced content generality and stylistic diversity for systematic evaluation of style transfer methods. Our main contributions are summarized as follows:

\begin{itemize}

\item We introduce a novel paradigm that reframes style transfer as a \textbf{supervised} learning problem via \textbf{destylization}. This reverse formulation enables unaltered style image to serve as direct learning targets, providing authentic supervision signals and effectively addressing the long-standing ``ground-truth'' absence problem.

\item We propose \textbf{DeStylePipe}, a progressive multi-stage destylization framework, tightly integrated with \textbf{DestyleCoT-Filter}, an interpretable evaluation mechanism that accurately assesses content preservation and style removal to guarantee stringent data quality.

\item We construct \textbf{DeStyle-350K}, which, to the best of our knowledge, is the largest-scale high-quality triplet dataset offering authentic supervision for style transfer. Comprising 350K high-quality triplets, it effectively overcomes the unreliability inherent in prior pseudo-target datasets.

\item We present \textbf{BCS-Bench}, a comprehensive benchmark featuring balanced  content generality and stylistic diversity . Utilizing this benchmark, we validate the high efficacy of DeStyle-350K by fine-tuning multiple models (e.g., Flux.2-Klein-9B/4B, Qwen-Image-Edit). The consistently superior results across different architectures firmly demonstrate that our dataset provides reliable and highly effective supervision for style transfer.

\end{itemize}
\section{Related Work}
\label{sec:related}

\noindent\textbf{Style Transfer.} 
Style transfer has evolved from early handcrafted filters~\cite{zhang2013style, wang2004efficient} and optimization-based feature matching~\cite{gatys2016image, gatys2017controlling, kolkin2019style} to efficient feed-forward models~\cite{huang2017arbitrary, li2017universal, liao2017visual, zhang2022exact, deng2020arbitrary}. Recently, diffusion-based methods~\cite{wang2024instantstyle, chung2024style, xu2024freetuner, xing2024csgo} have shown immense promise through both tuning-based~\cite{zhang2023inversion, zhang2023prospect, wang2023stylediffusion} and tuning-free~\cite{wang2024instantstyleplus, gao2024styleshot, qi2024deadiff} strategies. However, the persistent absence of definitive targets often forces these methods to rely on noisy learning signals, such as handcrafted statistical metrics or synthetic pseudo-supervision~\cite{wang2025omnistyle}. To address this fundamental limitation, we propose a destylization paradigm that extracts structure-aligned natural content directly from artistic images, thereby constructing grounded supervision pairs. Based on this, we introduce DeStyle-350K, providing authentic ground truth to reliably train and evaluate state-of-the-art style transfer models.

\noindent\textbf{Datasets for Style Transfer.} 
Early style-related datasets including WikiArt~\cite{artgan2018} and Style30K~\cite{li2024styletokenizer}, provide abundant artistic exemplars but lack the aligned triplets necessary for supervised training. While recent efforts have introduced synthetic triplet datasets~\cite{xing2024csgo, wang2025omnistyle}, they are inherently bottlenecked by the biases of the style transfer models used to create them. Furthermore, forcing MLLMs to evaluate these stylized domains often yields unreliable filtering, leading to noisy supervision and style drift. Our destylization-based pipeline circumvents this by reversing the stylization process to recover natural content, which aligns seamlessly with the primary training distribution of MLLMs. This approach enables robust, interpretable filtering, ultimately yielding a high-quality triplet dataset with accurate content alignment and authentic style supervision.

\section{DeStyle-350K Construction}
\label{sec:dataset}

\begin{figure*}[t]
    \centering
    \includegraphics[width=1.0\linewidth]
    {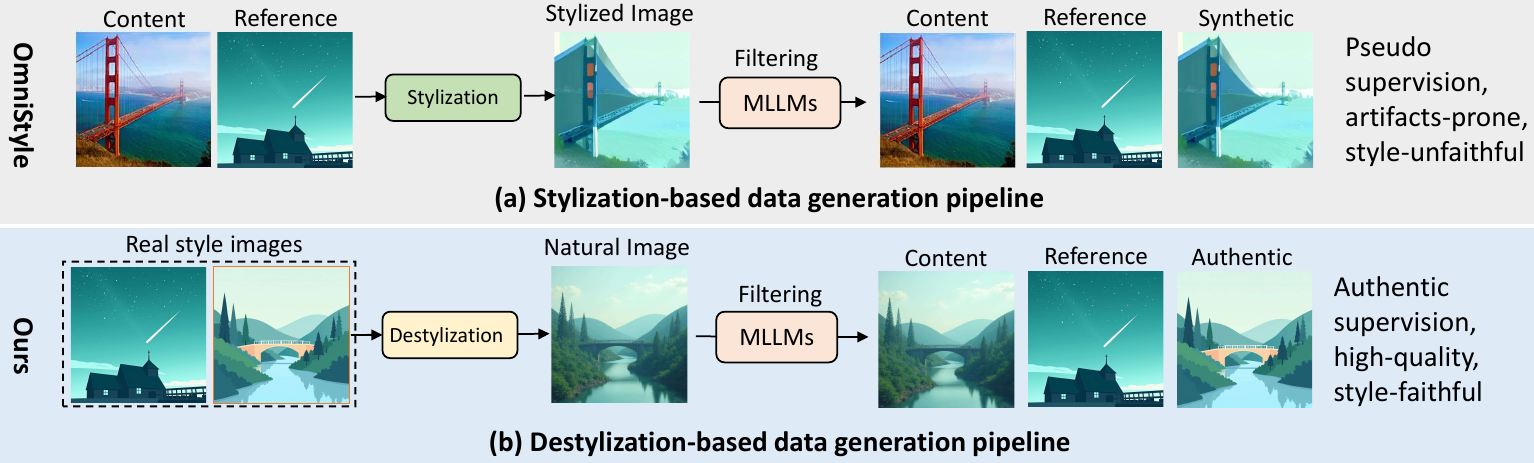}
    % \vspace{-2em}
    \caption{Stylization-based vs.\ destylization-based data generation pipelines.}
    \label{fig:desty_vs_sty}
    % \vspace{-1.5em}
\end{figure*}

\subsection{Motivation: From Stylization to Destylization}
\label{subsec:motivation}

A fundamental challenge in artistic style transfer is the absence of ground-truth pairs consisting of natural images and corresponding high-quality artistic counterparts. As illustrated in Fig. \ref{fig:desty_vs_sty}(a), previous paradigms, such as OmniStyle \cite{wang2025omnistyle}, primarily rely on a stylization-based data pipeline. These methods utilize existing style transfer models to apply artistic textures to natural content images, creating ``pseudo-triplets.'' However, this approach inherently suffers from a supervision bottleneck: the training signals are bounded by the limitations of the surrogate models, often inheriting artifacts, content leakage, and a significant distribution shift from real-world art. Consequently, models trained on such data struggle to capture the authentic essence of diverse artistic styles.

To break this limitation, we propose a paradigm shift: \textbf{learning to stylize by learning to destylize}, as shown in Fig.~\ref{fig:desty_vs_sty}(b). Instead of synthesizing pseudo-artworks, we start from authentic, high-quality style images, including classical masterpieces and high-fidelity synthetic art, and aim to recover their underlying natural content. Formally, this ``destylization'' process can be defined as $I_{natural} = \mathcal{F}_{desty}(I_{style} \mid T)$, where $\mathcal{F}_{desty}$ denotes our progressive destylization pipeline and $T$ represents the textual instructions that guide the model to reduce stylistic elements while preserving content structures. By systematically reversing the stylistic process, 
our framework provides \textbf{authentic style supervision} because the training target is the \textbf{unmodified style image}, be it a classical masterpiece or a high-fidelity synthetic artwork characterized by \textbf{integral artistic distribution}. Unlike stylization-based pipelines that rely on algorithmic approximations of a transfer process, our target images possess the coherent textures and global stylistic consistency of original art. Thus, the model learns from \textbf{primary stylistic distributions} rather than degraded, transformation-induced representations.

% \subsection{T、subsehe DeStyle-350K Dataset Construction}
% This section details the construction of DeStyle-350K, a large-scale dataset synthesized through our progressive destylization framework. By reversing the stylistic process, we generate 350K high-quality triplets, comprising 150K triplets from real-world artworks and 200K triplets from AI-generated style images. 

\subsection{Multi-source Artistic Image Collection}

\begin{figure*}[t]
    \centering
    \includegraphics[width=1.0\linewidth]
    {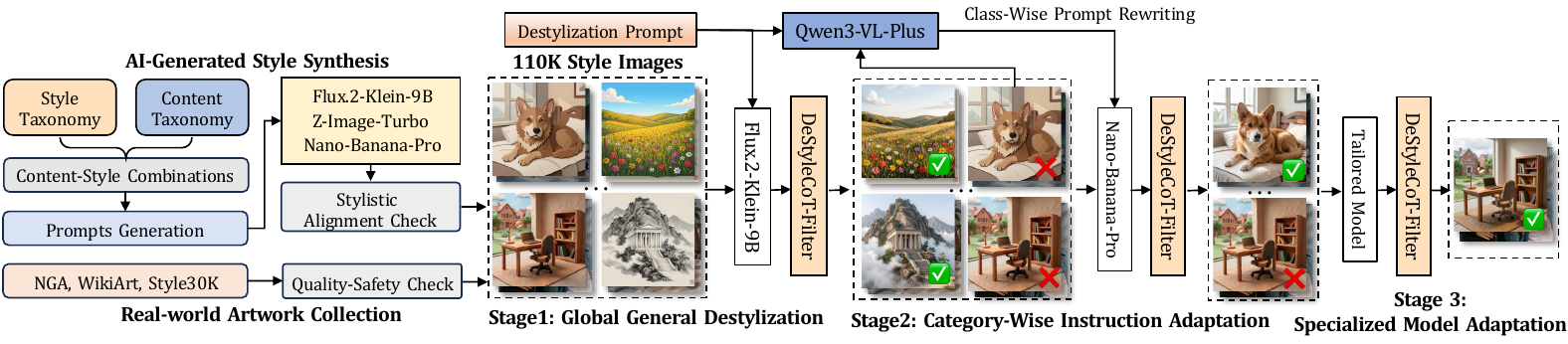}
    % \vspace{-2em}
    \caption{\textbf{Overview of DeStylePipe framework.} Starting from multi-source artistic image collection, a three-stage progressive destylization pipeline is applied, with DestyleCoT-Filter after each stage to assess quality and route failed cases forward.}
    \label{fig:dataset_construction}
    % \vspace{-2em}
\end{figure*}

\noindent\textbf{Real-world Artwork Collection.} We aggregate an extensive corpus of artistic images from diverse digital galleries (see Fig.\ref{fig:dataset_construction}), including the National Gallery of Art (NGA), WikiArt \cite{artgan2018}, and Style30K \cite{li2024styletokenizer}. To ensure quality and safety, we filter the collection for inappropriate content and unethical risks, followed by a resolution-based screening that discards samples below $1024 \times 1024$ pixels. This process yields a high-quality subset of 50K artworks spanning primary stylistic distributions across classical and modern human-crafted art.

\noindent \textbf{AI-Generated Style Synthesis. }
To expand stylistic diversity, we develop an automated generation pipeline based on our hierarchical taxonomy (see Fig.\ref{fig:dataset_construction}). First, we pair the 6 major artistic categories (abstract art, cultural regional, digital style, illustration and comic, material crafts, and traditional fine arts) with the 7 primary content categories (animals, architecture, human, interior, landscape, plant, and product and object) to create an exhaustive set of content-style combinations. Qwen3 \cite{qwen3} then expands these pairs into descriptive prompts to drive multiple T2I models, including Flux.2-Klein-9B \cite{flux-2-2025}, Z-Image-Turbo \cite{team2025zimage}, and Nano-Banana-Pro \cite{team2023gemini}, synthesizing 90K raw artistic images. Finally, to guarantee that each image strictly adheres to its intended style, we employ Qwen3-VL-Plus \cite{bai2023qwen} for a rigorous stylistic alignment check. Only samples that faithfully represent the target style are retained, resulting in approximately 60K high-fidelity images for subsequent destylization. A detailed taxonomy including all sub-categories is provided in the Supplementary Material.

\subsection{DeStylePipe: Progressive Multi-Stage Destylization Framework} 
Given the curated 110K artistic images, we introduce DeStylePipe, a progressive multi-stage framework designed to perform systematic destylization and recover the underlying natural content. The overall architecture is illustrated in Fig.~\ref{fig:dataset_construction} .

\noindent\textbf{Stage 1: Global General Destylization.} 
In the first stage, all 110K stylistic images undergo a foundational destylization. We leverage Flux.2-Klein-9B \cite{flux-2-2025} to perform the initial inference, guided by a universal instruction: \textit{“Transform this image into a style-free natural image.”}  This process efficiently handles samples with low stylistic complexity. Then, each output is evaluated by our DestyleCoT-Filter (Sec.~\ref{subsec:dst-filter}) based on content preservation and stylistic removal. Samples that satisfy these dual constraints are directly incorporated into the final dataset, while the remainder proceed to the subsequent stages for further processing.

\noindent\textbf{Stage 2: Category-Wise Instruction Adaptation.}
For samples failing Stage 1, we implement a more granular destylization strategy. First, Qwen3-VL-Plus \cite{bai2023qwen} is utilized to analyze these failed stylistic images, identifying their specific stylistic attributes to perform category-wise prompt rewriting. Then, we employ Nano-Banana-Pro \cite{team2023gemini} to execute the destylization with these tailored instructions. These customized prompts, coupled with the more advanced image editing model, enable more precise destylization in complex artistic domains. Finally, the outputs are re-evaluated by the DestyleCoT-Filter; samples satisfying the quality constraints are incorporated into the dataset, while the remaining failed cases proceed to Stage 3 for final processing.

\noindent \textbf{Stage 3: Specialized Model Adaptation.}
While prompt engineering works for most style domains, certain complex styles that are difficult to describe with text---such as needle felted, clay, or origami---remain hard to process even with tailored instructions. To solve this, we implement a specialized model adaptation strategy for these challenging cases. We fine-tune a customized destylization model via LoRA to specifically learn these complex style removals. The training dataset is constructed by sampling natural images from the HQ-50K\cite{yang2023hq} collection as ground-truth, and then using Nano-Banana-Pro to synthesize their stylized versions using these persistent styles as references. This process yields 20K structure-aligned pairs for direct supervision. Finally, after a validation by the DestyleCoT-Filter, these qualified samples are integrated in the final dataset.

\begin{figure*}[t]
    \centering
    \includegraphics[width=1.0\linewidth]
    {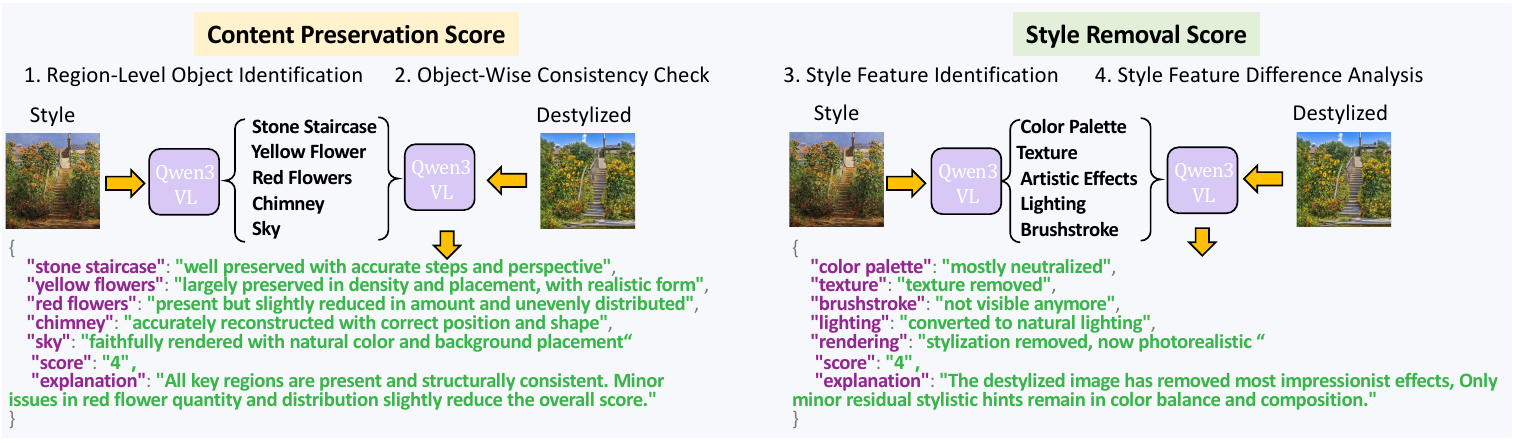}
    % \vspace{-2em}
    \caption{\textbf{Overview of DestyleCoT-Filter.} A CoT-driven multi-dimensional filtering framework scoring across content preservation and style removal to ensure data quality.}
    % \vspace{-2em}
    \label{fig:cot_filter}
\end{figure*}

% \vspace{-1em}
\subsection{DestyleCoT-Filter}
\label{subsec:dst-filter}

To ensure high data quality, we introduce DestyleCoT-Filter, a Chain-of-Thought-based mechanism designed to rigorously evaluate style-destylized image pairs. Unlike existing filters~\cite{wang2025omnistyle} that assess complex artistic domains, our approach evaluates the destylized outputs to better align with the pre-training data distribution of MLLMs. By shifting the assessment from abstract artistic styles to natural realism, we leverage the model's inherent understanding of natural scenes, leading to more stable and objective filtering results. As shown in Fig.~\ref{fig:cot_filter}, the pipeline evaluates two key dimensions through a structured reasoning process: content preservation and style removal.

\noindent \textbf{Content Preservation.}
Directly prompting MLLMs for consistency scores often overlooks fine-grained semantic drift. To mitigate this, we employ Qwen3-VL-Plus with a structured CoT strategy that decomposes the evaluation into three sequential steps. First, the model performs region-level object identification to extract key semantic components from the source style image, such as "stone staircase" or "chimney." Next, it executes an object-wise consistency check by cross-referencing these objects within the destylized output to verify their presence, placement, and structural integrity. Finally, the model provides a content preservation scoring (0-5) based on these specific assessments. This step-by-step reasoning grounds the final judgment in explicit visual evidence, capturing even minor discrepancies in object distribution.

\noindent \textbf{Style Removal.}
To quantify destylization degree, we adopt an attribute-based evaluation strategy consisting of two main steps. First, the model performs style feature identification to decompose the original image into distinct style dimensions, such as color palette, texture, brushstrokes, and lighting. Next, it executes a style feature difference analysis to evaluate the ``residual style'' in the destylized output. During this stage, the model checks if artistic elements have been neutralized or converted to natural standards, such as verifying if textures are removed or brushstrokes are no longer visible. Finally, the model assigns a style difference score (0-5) with a formal explanation. This structured reasoning ensures that the assessment captures specific stylistic remnants across all identified dimensions, leading to an objective measure of successful destylization.

\noindent \textbf{Filtering Criterion.}
To ensure high data quality, we implement a dual-threshold selection strategy. A sample is admitted to the final dataset only if it achieves a score of $\ge$ 4 in both content preservation and style removal. This strict rule ensures that every pair is structurally consistent with the source and free of stylistic artifacts. By enforcing these constraints, we provide high-quality data for training robust stylization models.

\noindent \textbf{Triplet Construction.}
Based on the high-quality samples filtered above, we construct triplets to facilitate stable training and prevent semantic leakage. Each triplet consists of a destylized image (content), a style reference (condition), and a style image (target). For each style target, we retrieve the Top-5 most similar style references from different semantic categories (e.g., matching an ``animal'' target with ``non-animal'' references). For AI-generated data, categories are derived from our content taxonomy, while for real-world art, they are identified by Qwen3-VL-Plus. If fewer than five candidates exist, we retain the maximum available. This cross-semantic matching ensures the model decouples style from category-specific priors, significantly enhancing training robustness. This rigorous triplet construction process yields the DeStyle-350K dataset.

\begin{table}[t]
\centering
\renewcommand{\arraystretch}{0.6}
\setlength{\tabcolsep}{32pt}
\caption{Evaluation of destylization quality. All scores are
averaged on a 0-5 scale (higher is better).}
% \vspace{-1em}
\label{tab:quantitative_results}
\resizebox{1.0\textwidth}{!}{
\begin{tabular}{lccc}
\toprule
\textbf{Evaluation Setting} & \textbf{Criterion} & \textbf{Scope} & \textbf{Avg. Score} \\
\midrule
\multirow{2}{*}{DeStyleCoT-Filter} 
& Style Removal & DeStyle-350K & 4.57 \\
& Content Preservation & DeStyle-350K & 4.64 \\
\midrule
\multirow{3}{*}{Expert User Study} 
& Realism & 1,000 Triplets & 4.52 \\
& Content Preservation & 1,000 Triplets & 4.71 \\
& Style Removal & 1,000 Triplets & 4.48 \\
\bottomrule
% \vspace{-3.5em}
\end{tabular}}
\end{table}

\noindent\textbf{Destylization Evaluation.}
As shown in the upper section of Table~\ref{tab:quantitative_results}, 
we first apply the proposed DeStyleCoT-Filter to evaluate the entire DeStyle-350K dataset along two key dimensions. 
Across this large-scale dataset, DeStylePipe achieves an average style removal score of 4.57 and a content preservation score of 4.67 on a 5-point scale. 
These results indicate that our pipeline effectively removes artistic attributes while preserving structural consistency. Furthermore, as shown in the lower section of Table~\ref{tab:quantitative_results}, 
we further conduct a comprehensive user study involving 10 experts with artistic backgrounds. 
The experts evaluate 1,000 randomly sampled triplets, each consisting of a de-stylized output, the reference content image, and the corresponding style image. 
For each triplet, scores ranging from 0 to 5 are assigned across three criteria: 
(1) \textit{Realism}, 
(2) \textit{Structural Preservation}, and 
(3) \textit{Style Removal}. 
Although the absolute scores differ, the expert evaluation exhibits a similar performance trend, further corroborating the effectiveness of DeStylePipe.

% \begin{figure*}[t]
%     \centering
%     \includegraphics[width=1.0\linewidth]
%     {figures_new/multi_ref.pdf}
%     \caption{
%        (a) Destylization Dataset Construction and 
%        (b) The architecture of DestyleNet model.
%     }
%     \label{fig:cot_filter}
% \end{figure*}
% \vspace{-1em}
\subsection{Dataset Statistics}

\begin{figure*}[t]
    \centering
    \includegraphics[width=\linewidth]{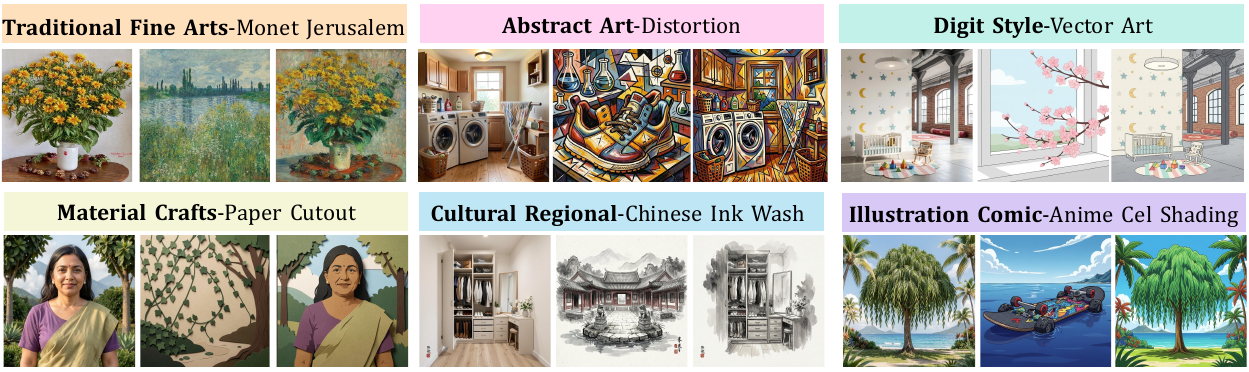}
    % \vspace{-2em}
    \caption{\textbf{Representative samples of DeStyle-350K.} Our dataset spans six major style categories with over 500 fine-grained subcategories. Each triplet shows (left to right) the destylized image, reference image, and style image.}
    \label{fig:dataset}
    % \vspace{-1em}
\end{figure*}

\noindent \textbf{Overview.}
DeStyle-350K is a large-scale dataset consisting of 350K high-quality training triplets. This dataset comprises 150K samples derived from real-world artistic images and 200K samples from AI-generated stylistic images. As illustrated in Fig.~\ref{fig:dataset}, the dataset provides extensive coverage across 6 major artistic categories and 7 primary content categories. Specifically, DeStyle-350K encompasses over 500 fine-grained stylistic variations (e.g., \textit{Ukiyo-e}, \textit{Cyberpunk}, and \textit{Needle Felted}) and 100 content classes, ensuring both stylistic diversity and content richness. More detailed information is provided in the Appendix.

\noindent \textbf{Comparative Analysis.}
We compare DeStyle-350K with OmniStyle150K, the current largest open-source style transfer dataset, as summarized in Table~\ref{tab:dataset_comparison}. Our dataset provides a substantial expansion in scale (350K vs. 150K) and introduces a hybrid data source strategy by integrating real-world artistic images with synthetic ones. In terms of diversity, DeStyle-350K covers over 100 content categories and 30K style references, offering significantly finer granularity compared to the 20 categories and 1,000 style references in OmniStyle150K. Quantitative metrics for style consistency, content preservation, and visual quality are obtained by averaging Qwen3-VL-Plus scores across the entire dataset. These metrics are measured on a scale of 0-5 (higher is better). DeStyle-350K consistently achieves higher scores in all dimensions, particularly in style consistency score (4.54 vs. 3.78), indicating that target images synthesized by conventional style transfer models often suffer from style artifacts. Further details regarding the scoring prompts and evaluation protocols are provided in the Appendix.

\begin{table}[t]
\centering
\caption{\textbf{Comparison between OmniStyle150K and our DeStyle-350K.} Our dataset provides significantly larger scale, superior diversity, and enhanced quality.}
% \vspace{-1em}
\label{tab:dataset_comparison}
\setlength{\tabcolsep}{8pt}
\renewcommand{\arraystretch}{1.2}
\definecolor{highlightpurple}{RGB}{240, 230, 255} % 定义浅紫色
\resizebox{\linewidth}{!}{
\begin{tabular}{lccccccc}
\toprule
Dataset & Size & Content Cat. & Style Ref. & Supervision & Style Cons. $\uparrow$ & Cont. Pres. $\uparrow$ & Vis. Qual. $\uparrow$ \\
\midrule
OmniStyle150K & 150K & 20 & 1K ref & Synthetic & 3.78 & 4.46 & 4.24 \\
\rowcolor{highlightpurple} \textbf{DeStyle-350K} & \textbf{350K} & \textbf{100+} & \textbf{30K ref} & \textbf{Real} & \textbf{4.54} & \textbf{4.67} & \textbf{4.71} \\
\bottomrule
% \vspace{-4em}
\end{tabular}
}
\end{table}

\section{Experiments}
\label{sec:experiment}
% \vspace{-1em}
\subsection{BCS-Bench and Experimental Setup}
\label{subsec:setup}

\begin{figure*}[t]
    \centering
    \includegraphics[width=\linewidth]{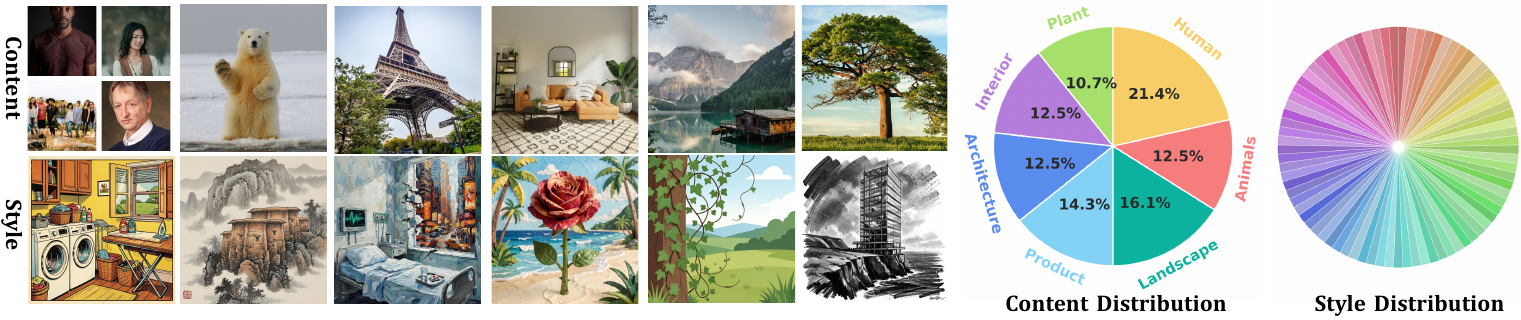}
    % \vspace{-2em}
    \caption{
    Content and style examples (left) and distributions (right) of BCS-Bench. For clarity, 81 style categories information are omitted.
    }
    \label{fig:bench}
    % \vspace{-1.5em}
\end{figure*}

\noindent\textbf{BCS-Bench.}
As summarized in Table~\ref{tab:benchmark_comparison}, prior benchmarks primarily focus on style diversity while overlooking content variation: most use small, fixed content sets (e.g., 20 images) with limited semantic breadth and demographic balance, and some employ stylized or non-natural images, weakening their value as structural references. To address this gap, we introduce BCS-Bench (Fig.~\ref{fig:bench}), a curated benchmark that balances content generality with stylistic diversity. It comprises 81 style images spanning 81 artistic styles (from 2D pixel art/sketch to 3D origami/clay art) paired with 60 content images across 7 major categories (animals, architecture, human, interior, landscape, plant, and product). Taking Human as an example (Fig.~\ref{fig:bench}, upper-left), we stratify by age, skin tone, body shape, disability, and group composition (single vs. multi-person) for fair, diversity-aware evaluation. These pairings yield 4,860 unique combinations at $1024\times1024$, enabling comprehensive quantitative and qualitative evaluation.

\noindent\textbf{Baselines.}
We evaluate our approach against two groups of methods: (1) Style Transfer Models: OmniStyle \cite{wang2025omnistyle}, USO \cite{wu2025uso},Qwen-Style \cite{zhang2026qwenstyle}, Attention Distillation (AD) \cite{zhou2025attention}, StyleID \cite{chung2024style}, CSGO \cite{xing2024csgo}, StyleShot \cite{gao2024styleshot}. (2) Image Editing Models: Closed- and open-source models including Nano-Banana-Pro, GPT-Image-1.5, Qwen-Image-Edit~\cite{wu2025qwen}, FLUX.2-Klein9B~\cite{flux-2-2025}, FLUX.2-Klein4B~\cite{flux-2-2025}.

\begin{table}[t]
\centering
\small
\caption{Comparison of existing style transfer benchmarks and our proposed BCS-Bench. ``N/A'' denotes missing information.}
% \vspace{-1em}
\label{tab:benchmark_comparison}
\definecolor{highlightpurple}{RGB}{240, 230, 255} % 定义浅紫色
\resizebox{\linewidth}{!}{
\begin{tabular}{lcccccc}
\toprule
\textbf{Benchmark} & \textbf{Content Images} & \textbf{Content Categories} & \textbf{Content Ethics} & \textbf{Style Categories} & \textbf{Content-Style Pairs} & \textbf{Resolution} \\
\midrule
CAST \cite{zhang2020cast}             & N/A & N/A & N/A & N/A & 50   & N/A \\
AesPANet \cite{hong2023aespa}         & N/A & N/A & N/A & N/A & 65   & 256$\times$256 \\
InST \cite{zhang2023inversion}             & N/A & N/A & N/A & N/A & 26   & N/A \\
StyleID \cite{chung2024style}          & 20 & 4 & No  & Only Oil paintings & 800  & 512$\times$512 \\
StyleShot \cite{gao2024styleshot}        & 20  & 6  & No & 73  & 9,800 & 879$\times$876 \\
OmniStyle \cite{wang2025omnistyle}        & 20  & 4  & No & 32 & 2,000 & 1024$\times$1024 \\
% \midrule
\rowcolor{highlightpurple} \textbf{BCS-Bench (Ours)} & \textbf{60} & \textbf{7} & Yes & \textbf{81} & \textbf{4,860} & \textbf{1024}$\times$\textbf{1024} \\
\bottomrule
\end{tabular}
% \vspace{-1em}
}
\end{table}

\noindent\textbf{Evaluation Metrics.}
Our model is evaluated across three dimensions: content preservation (DINO similarity, CLIP text-image alignment), style similarity (CSD score~\cite{somepalli2024measuring}, style loss), and VLM-based assessment. For the choice of VLM, we choose Qwen3-VL-Plus~\cite{bai2023qwen} to score four aspects from 0 to 5: Qwen-C (content preservation), Qwen-S (style similarity), Qwen-A (aesthetic quality), and Qwen-L (style semantic leakage, lower is better). Qwen-L measures how much semantic content from the style reference erroneously appears in the stylized output; lower indicates better style-content disentanglement.

\noindent \textbf{Implementation Details.}
We train three baseline models on the DeStyle-350K dataset: Flux.2-Klein-9B, Flux.2-Klein-4B, and Qwen-Image-Edit-2511. For the Flux-based models, we perform full fine-tuning with a learning rate of 1e-5 and a batch size of 32. For Qwen-Image-Edit-2511, we use LoRA fine-tuning (lr=1e-4, batch size 8) due to its large parameter count. All experiments are conducted on a cluster of 8$\times$H20 GPUs.

% \vspace{-1em}
\subsection{Validation of Destylization Quality}
\label{subsec:destylization_quality}

\begin{figure*}[t]
    \centering
    \includegraphics[width=\linewidth]{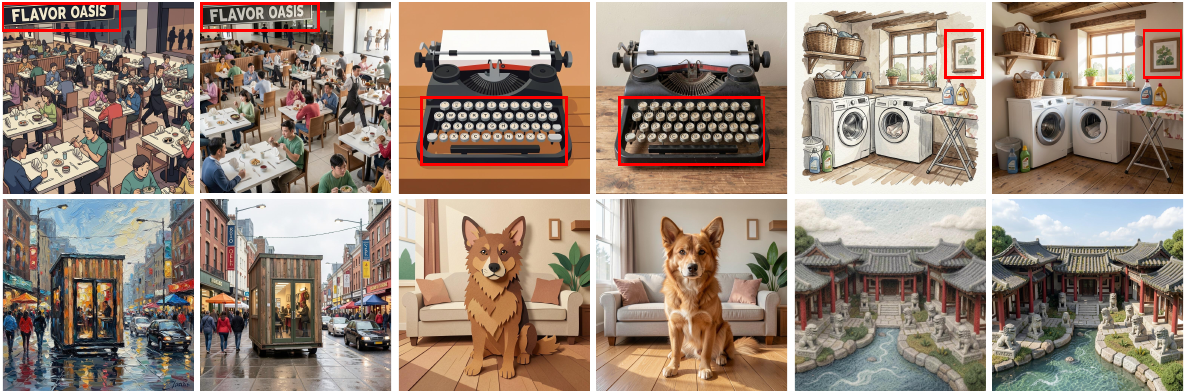}
    % \vspace{-2em}
    \caption{\textbf{Our destylization results.} Red bounding boxes highlight the preservation of fine-grained details and spatial structures when destylization, including text, keyboard keys, and intricate textures, without geometric distortion.}
    \label{fig:desty_qualitative}
    
    \includegraphics[width=1.0\linewidth]{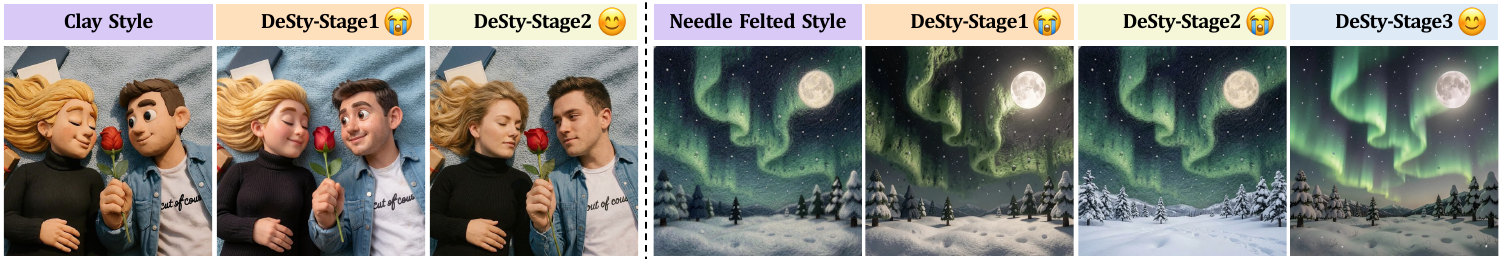}
    % \vspace{-2em}
    \caption{Destylization performance across stages in challenging styles.}
    \label{fig:desty_ablation}
    % \vspace{-2em}
\end{figure*}

\noindent\textbf{Qualitative Destylization Results.} As illustrated in Fig. \ref{fig:desty_qualitative}, our DeStylePipe globally processes diverse artistic inputs and successfully translates them into highly realistic natural images. It effectively removes stylistic artifacts across various domains, including but not limited to dense oil paintings, flat 2D illustrations, and 3D papercrafts, yielding convincing photorealistic appearances. Beyond this authentic global destylization, our method excels in preserving intricate local structures. As highlighted by the red bounding boxes, DeStylePipe maintains near pixel-perfect alignment for complex local geometries, flawlessly retaining legible textual elements like the ``FLAVOR OASIS'' sign, highly structured objects such as the typewriter keyboard, and fine background semantics like the wall-mounted picture frame. This precise structural preservation, coupled with effective style elimination, ensures that our extracted natural images provide pixel-aligned content inputs for style transfer while seamlessly aligning with the real-world data distribution. Fig.~\ref{fig:desty_ablation} further illustrates how DeStylePipe handles challenging styles through progressive refinement. For Clay Style and Needle Felted Style, Stage~1 yields unsatisfactory results with residual artifacts, while later stages progressively remove stylistic elements and produce photorealistic outputs. In the Needle Felted example, Stage~3 is required to fully recover the natural scene, showing that difficult styles benefit from the multi-stage pipeline.

% \begin{figure*}[t]
%     \centering
%     \includegraphics[width=\linewidth]{figures_new/desty_dingxing.pdf}
%     \vspace{-2em}
%     \caption{\textbf{Our destylization results.} Red bounding boxes highlight the preservation of fine-grained details and spatial structures when destylization, including text, keyboard keys, and intricate textures, without geometric distortion.}
%     \vspace{-1em}
%     \label{fig:desty_qualitative}
% \end{figure*}

% \begin{figure*}[!t]
%     \centering
%     \includegraphics[width=1.0\linewidth]
%     {figures_new/desty_ablation_dingxing.pdf}
%     \vspace{-2em}
%     \caption{Destylization performance across stages in challenging styles.}
%     \label{fig:desty_ablation}
%     \vspace{-2em}
% \end{figure*}

% \begin{table}[t]
% \centering
% \setlength{\tabcolsep}{18pt}
% \caption{Quantitative evaluation of data quality on DeStyle-350K. All scores are averaged on a 0-5 scale (higher is better).}
% \vspace{-1em}
% \label{tab:quantitative_results}
% \resizebox{1.0\textwidth}{!}{
% \begin{tabular}{lccc}
% \toprule
% \textbf{Evaluation Setting} & \textbf{Criterion} & \textbf{Scope} & \textbf{Avg. Score} \\
% \midrule
% \multirow{2}{*}{DeStyleCoT-Filter} 
% & Style Removal & DeStyle-350K & 4.57 \\
% & Content Preservation & DeStyle-350K & 4.64 \\
% \midrule
% \multirow{3}{*}{Expert User Study} 
% & Realism & 1,000 Triplets & 4.52 \\
% & Content Preservation & 1,000 Triplets & 4.71 \\
% & Style Removal & 1,000 Triplets & 4.48 \\
% \bottomrule
% \vspace{-2.5em}
% \end{tabular}}
% \end{table}

% \vspace{-1.5em}
\subsection{Validation of Stylization Performance}
% \vspace{-1em}
\label{subsec:sota_comparison}
Unless otherwise noted, all experimental evaluations and visualizations are based on the fine-tuned Flux.2-Klein-9B model.

\begin{figure*}[!h]
    \centering
    \includegraphics[width=1.0\linewidth]
    {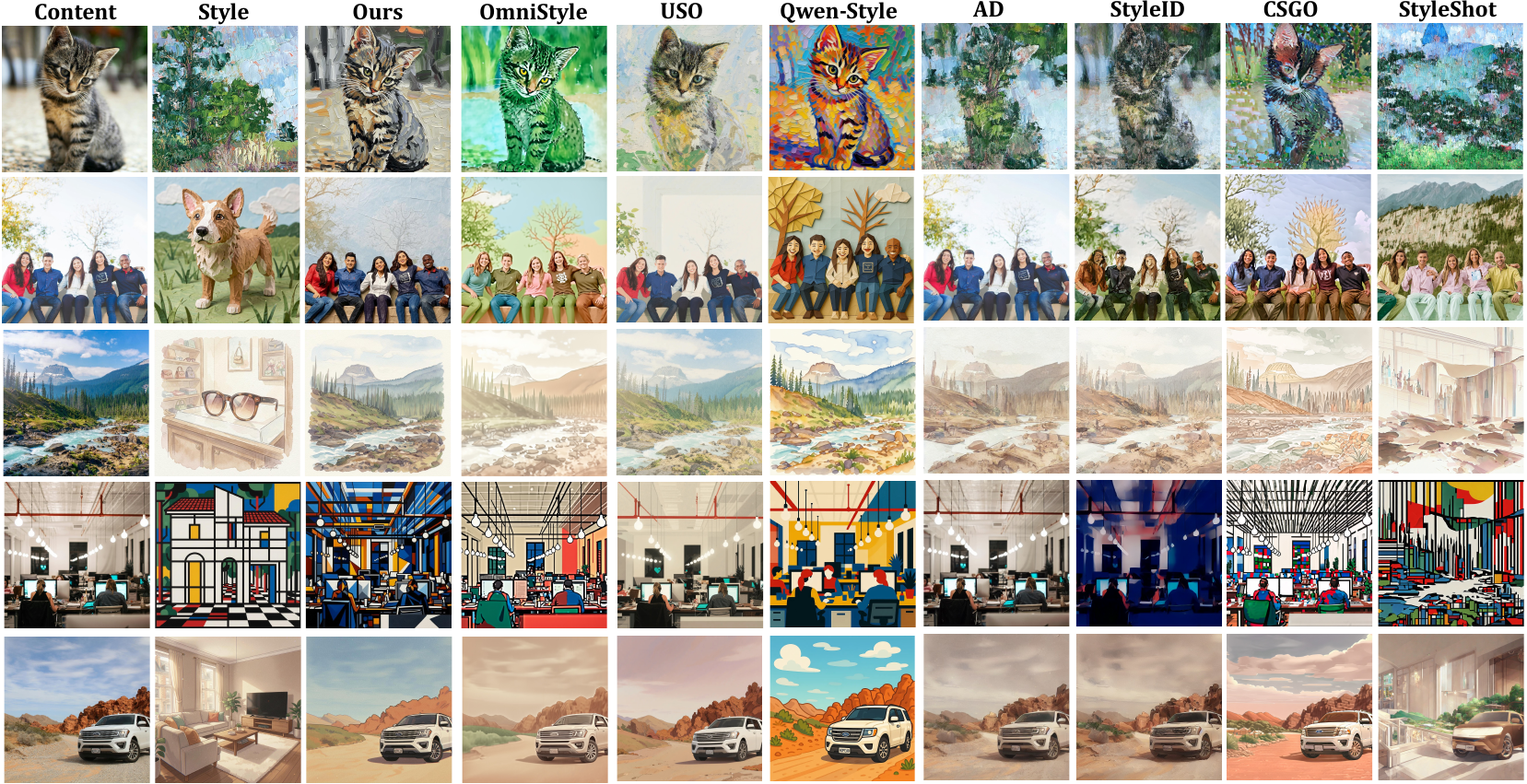}
    % \vspace{-2em}
    \caption{
       Qualitative comparison with other state-of-the-art style transfer methods. Note that our method keep the human skin tone and face identity without stylization distortions.
    }
    \label{fig:qualitative_compare1}
    % \vspace{-1em}
\end{figure*}

\noindent\textbf{Qualitative Comparison with Style Transfer Models.} As illustrated in Fig. \ref{fig:qualitative_compare1}, our method significantly outperforms state-of-the-art baselines in both stylistic expression and structural faithfulness. Regarding stylistic expression, baselines such as StyleShot and CSGO (Rows 1 and 3) suffer from ``content leakage'', where semantic elements from the style reference erroneously appear in the output, while USO (Rows 2 and 4) exhibits insufficient stylization for complex artistic style. In contrast, our model achieves a profound transformation, accurately capturing elements like thick brushstrokes and abstract geometric blocks. In terms of content preservation, our method effectively avoids the ``structural collapse'' or ``content hallucinations'' prevalent in Qwen-Style (Row 2 and 5), which often leads to distorted facial identities and inconsistent global scale. Notably, our approach maintains precise local details even in complex scenes (Rows 4 and 5); for instance, in Row 5, our result preserves consistent license plate information while all other methods fail. These advantages underscore the value of the DeStyle-350K dataset, whose high-fidelity, pixel-aligned nature provides the precise supervision necessary to overcome the traditional trade-off between style strength and content integrity.

\begin{figure*}[!h]
    \centering
    \includegraphics[width=1.0\linewidth]
    {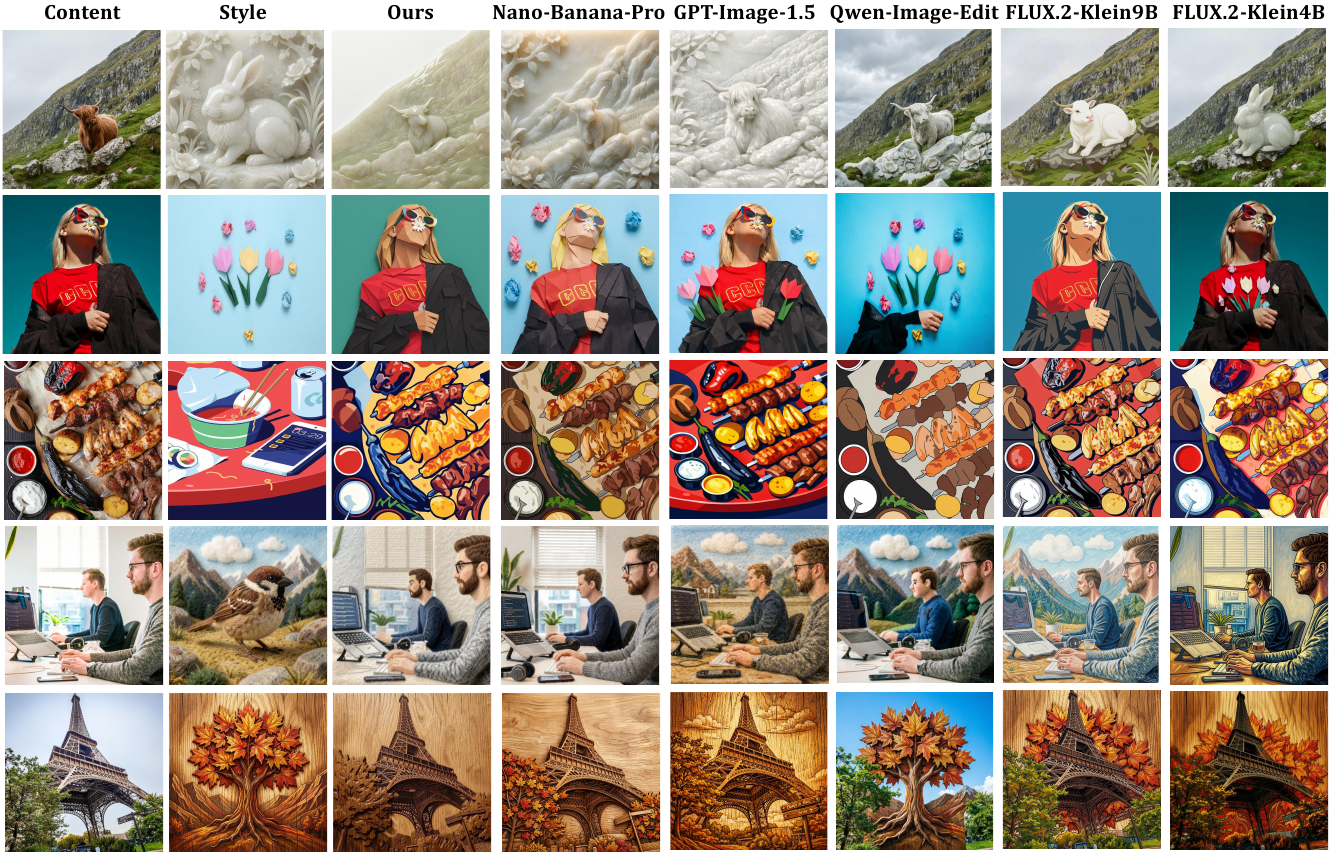}
    % \vspace{-2em}
    \caption{
       Qualitative comparison with the existing image editing models.
    }
    \label{fig:qualitative_compare2}
    % \vspace{-2em}
\end{figure*}

\noindent\textbf{Qualitative Comparison with Image Editing Models.} As shown in Fig. \ref{fig:qualitative_compare2}, we first examine representative open-source models, including Qwen-Image-Edit and the FLUX2 series, under complex style conditions. As shown in Rows 1, 2, 4, and 5, these models exhibit recurring failure modes. Qwen-Image-Edit frequently suffers from \textit{semantic leakage}, introducing unintended elements from the style reference. The FLUX2 variants demonstrate either incorrect style transfer (Row 1) or insufficient stylization (Rows 2 and 4), failing to faithfully capture complex stylistic attributes. We further analyze closed-source systems, Nano-Banana-Pro and GPT-Image-1.5. While GPT-Image-1.5 achieves stronger stylization, it often alters the original content structure and disrupts global spatial scale (Rows 1 and 3). Nano-Banana-Pro preserves structure more reliably but still exhibits semantic leakage under complex scenarios. In contrast, our method consistently preserves semantic content and geometric structure while accurately transferring high-order stylistic patterns, even under highly complex style (Row 4).

\noindent\textbf{Quantitative Comparison with Style Transfer Models.}
As shown in Table~\ref{tab:comprehensive_comparison}, our method ranks best on six of eight metrics, leading in both style fidelity (CSD=0.5485, Qwen-S=4.1140) and content preservation (CLIP-TI=0.3034, Qwen-C=4.5610), with the lowest semantic leakage (Qwen-L=0.0059). In contrast, USO and StyleID preserve content well (DINO$>$0.68) but transfer style poorly (Qwen-S$<$2.90), whereas StyleShot achieves low Style Loss at the expense of severe content degradation (DINO=0.2963). Therefore, our approach strikes the best balance across all evaluation dimensions.

\begin{table*}[t]
\centering
\small
\definecolor{highlightpurple}{RGB}{240, 230, 255} % 定义浅紫色
\setlength{\tabcolsep}{8pt}
\caption{Quantitative comparison of style transfer methods across multiple metrics (\textbf{best} in bold, \uline{second-best} underlined).}
% \vspace{-1em}
\label{tab:comprehensive_comparison}
\resizebox{1.0\textwidth}{!}{
\begin{tabular}{lcccccccc}
\toprule
\textbf{Method / Metric} & \textbf{DINO $\uparrow$} & \textbf{CLIP-TI $\uparrow$} & \textbf{CSD $\uparrow$} & \textbf{Style Loss $\downarrow$} & \textbf{Qwen-C $\uparrow$} & \textbf{Qwen-S $\uparrow$} & \textbf{Qwen-A $\uparrow$} & \textbf{Qwen-L $\downarrow$} \\
\midrule
OmniStyle   & 0.6579 & 0.2874& 0.4551 & 0.1047& 3.9131& 2.8840& 4.3089& 0.0135\\
USO         & \textbf{0.6871} & \uline{0.2975}& 0.3902 & 0.1106& \uline{4.4606}& 2.8612& \uline{4.4882}& 0.0098 \\
Qwen-Style  & 0.5877 & 0.2861& 0.5122 & 0.1152& 3.6781& 3.4156& 4.2265& 0.0765\\
AD          & 0.6111 & 0.2824& 0.4523 & 0.1221& 3.7790& 2.7014& 4.0808& 0.3494 \\
StyleID     & \uline{0.6860} & 0.2901& 0.3872 & 0.1194& 3.9272& 2.5330& 4.2663& \uline{0.0067}\\
CSGO        & 0.6022 & 0.2885& 0.4750 & 0.1854& 3.7027& \uline{3.3800}& 4.3867& 0.0208\\
StyleShot   & 0.2963 & 0.2009& \uline{0.5308} & \textbf{0.0833} & 2.0560&  3.0644& 4.4452& 0.0473\\
\rowcolor{highlightpurple}  Ours & 0.6441 & \textbf{0.3034}& \textbf{0.5485} & \uline{0.0916}& \textbf{4.5610}& \textbf{4.1140}& \textbf{4.5269}& \textbf{0.0059}\\
\bottomrule
\end{tabular}
}
\end{table*}

\begin{table*}[t]
\centering
\definecolor{highlightpurple}{RGB}{240, 230, 255} % 定义浅紫色
\small
\setlength{\tabcolsep}{8pt}
\caption{Quantitative comparison of image editing methods across multiple metrics (\textbf{best} in bold, \uline{second-best} underlined).}
% \vspace{-1em}
\label{tab:comparison_4o}
\resizebox{1.0\textwidth}{!}{
\begin{tabular}{lcccccccc}
\toprule
\textbf{Model / Metric} & \textbf{DINO $\uparrow$} & \textbf{CLIP-TI $\uparrow$} & \textbf{CSD $\uparrow$} & \textbf{Style Loss $\downarrow$} & \textbf{Qwen-C $\uparrow$} & \textbf{Qwen-S $\uparrow$} & \textbf{Qwen-A $\uparrow$} & \textbf{Qwen-L $\downarrow$} \\
\midrule
Qwen-Image-Edit  & 0.5248& 0.2337& 0.4967& 
0.1482& 3.2562& 3.6436& 4.5403& \uline{0.0088}\\
FLUX.2-Klein9B   & 0.5191 & 0.2523& \uline{0.6240} & 0.0562& 3.7121& 4.0858& 4.6108& 0.4748 \\
FLUX.2-Klein4B   & 0.4985& 0.2451&\textbf{0.6362}  & \uline{0.0504}& 4.0437& 3.5424& \uline{4.6138}& 0.4380 \\
GPT-Image-1.5    & \textbf{0.6506} & \uline{0.2952}& 0.5305 & 0.0892& \uline{4.5541}& \uline{4.4389}& \textbf{4.7143}& 0.0691\\
Nano-Banana-Pro  & 0.6047 & 0.2801& 0.5674 & \textbf{0.0434}& 4.5378& \textbf{4.4552}& 4.6092& 0.1185 \\
\rowcolor{highlightpurple} Ours             & \uline{0.6441} & \textbf{0.3034}& 0.5485 & 0.0916& \textbf{4.5610}& 4.1140& 4.5269& \textbf{0.0059} \\
\bottomrule
\end{tabular}
}
\end{table*}

\noindent\textbf{Quantitative Comparison with Image Editing Models.}
As shown in Table~\ref{tab:comparison_4o}, our method achieves the best scores in CLIP-TI, Qwen-C, and Qwen-L, and ranks second in DINO, indicating strong content preservation and style-content disentanglement. It lags behind general-purpose editing models (e.g., Nano-Banana-Pro, GPT-Image-1.5) on CSD and Qwen-S, as those models are optimized for broader editing tasks, whereas ours is tailored for style transfer.

\begin{table*}[t]
\centering
\small
\setlength{\tabcolsep}{10pt}
\caption{\textcolor{black}{User study comparison between our method and representative style transfer approaches (\textbf{best} in bold, \uline{second-best} underlined).}}
% \vspace{-1em}
\label{tab:us_1}
\resizebox{1.0\textwidth}{!}{
\begin{tabular}{lcccccccc}
\toprule
\textbf{Metric / Method} & \textbf{Ours} & \textbf{OmniStyle} & \textbf{USO} & \textbf{Qwen-Style} & \textbf{AD} & \textbf{StyleID} & \textbf{CSGO} & \textbf{StyleShot} \\
\midrule
Rank 1 (\%) $\uparrow$ & \textbf{28.19} & 10.82 & 9.65 & \uline{13.68} & 9.17 & 11.19 & 8.12 & 9.18 \\
Top 3 (\%) $\uparrow$ & \textbf{56.65} & 50.45 & 24.11 & \uline{51.27} & 31.25 & 33.68 & 25.58 & 27.01 \\
\bottomrule
% \vspace{-2em}
\end{tabular}
}
\end{table*}

\begin{table*}[t]
\centering
\small
\caption{\textcolor{black}{User study comparison between our method and representative image editing methods (\textbf{best} in bold, \uline{second-best} underlined).}}
% \vspace{-1em}
\label{tab:us_2}
\resizebox{1.0\textwidth}{!}{
\begin{tabular}{lcccccc}
\toprule
\textbf{Metrics/Model} & \textbf{Ours} & \textbf{Nano-Banana-Pro} & \textbf{GPT-Image-1.5} & \textbf{Qwen-Image-Edit} & \textbf{FLUX.2-Klein9B} & \textbf{FLUX.2-Klein4B} \\
\midrule
Rank 1 (\%) $\uparrow$ & 22.56 & \textbf{24.64} & \uline{22.92} & 11.96 & 12.55 & 5.37 \\
Top 3 (\%) $\uparrow$ & \uline{68.60} & \textbf{70.12} & 67.18 & 47.53 & 34.56 & 12.01 \\
\bottomrule
\end{tabular}
}
% \vspace{-1em}
\end{table*}

\begin{table*}[!t]
\centering
\caption{Multiple baselines' performance gains from fine-tuning on DeStyle-350K.}
% \vspace{-1em}
\label{tab:quantitative_results_backbone}
\definecolor{highlightpurple}{RGB}{240, 230, 255} % 定义浅紫色
\setlength{\tabcolsep}{4pt}
\resizebox{\linewidth}{!}{
\begin{tabular}{llcccccccc}
\toprule
\textbf{Backbone} & \textbf{Setting} 
& \textbf{DINO $\uparrow$} & \textbf{CLIP-TI $\uparrow$} & \textbf{CSD $\uparrow$} & \textbf{Style Loss $\downarrow$} 
& \textbf{Qwen-C $\uparrow$} & \textbf{Qwen-S $\uparrow$} & \textbf{Qwen-A $\uparrow$} & \textbf{Qwen-L $\downarrow$} \\
\midrule

\multirow{2}{*}{Flux.2-Klein4B}
& Before Finetuning 
& 0.4985& 0.2451&\textbf{0.6362}  & \textbf{0.0504}& 4.0437& \textbf{3.5424}& 4.4138& 0.4380 \\

& After Finetuning 
& \textbf{0.6930}& \textbf{0.2987}& 0.4999& 0.1085& \textbf{4.5286}& 3.1937& \textbf{4.4693}& \textbf{0.0025} \\

\midrule

\multirow{2}{*}{Flux.2-Klein9B}
& Before Finetuning 
& 0.5191& 0.2523& \textbf{0.6240}& 
\textbf{0.0562}& 3.7121& 4.0858& 4.4108& 0.4748 \\

& After Finetuning 
& \textbf{0.6441} & \textbf{0.3034}& 0.5485 & 0.0916& \textbf{4.5610}& \textbf{4.1140}& \textbf{4.5269}& \textbf{0.0059}\\

\midrule

\multirow{2}{*}{Qwen-Image-Edit}
& Before Finetuning 
& 0.5248& 0.2337& 0.4967& 
0.1482& 3.2562& 3.6436& \textbf{4.5403}& 0.0088\\

& After Finetuning (LoRA) 
& \textbf{0.6365}& \textbf{0.3059}& \textbf{0.5248}& 
\textbf{0.1105}& \textbf{4.5013}& \textbf{3.8954}& 4.4982& \textbf{0.0054}\\

\bottomrule
\end{tabular}
}
\end{table*}

\begin{figure*}[!t]
    \centering
\includegraphics[width=1.0\linewidth]
    {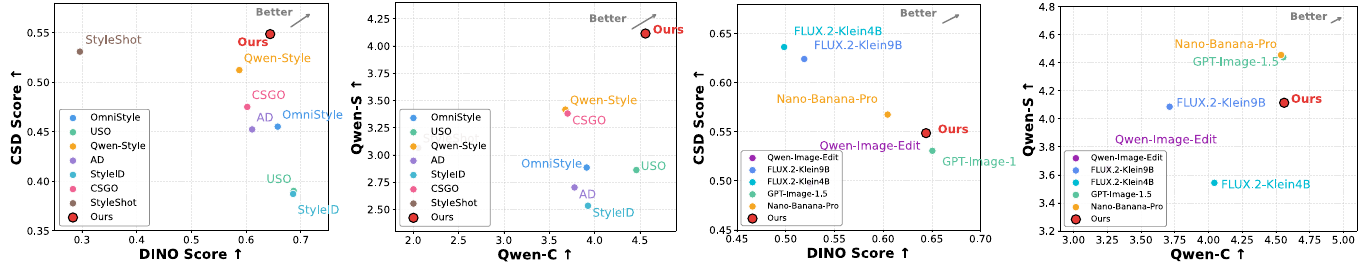}
    % \vspace{-2em}
    \caption{Trade-off between content preservation and style fidelity across style transfer (left two) and image editing (right two) methods. Upper-right is better.}
    \label{fig:scatter}
    % \vspace{-1em}
\end{figure*}

\begin{figure*}[!t]
    \centering
    \includegraphics[width=1.0\linewidth]
    {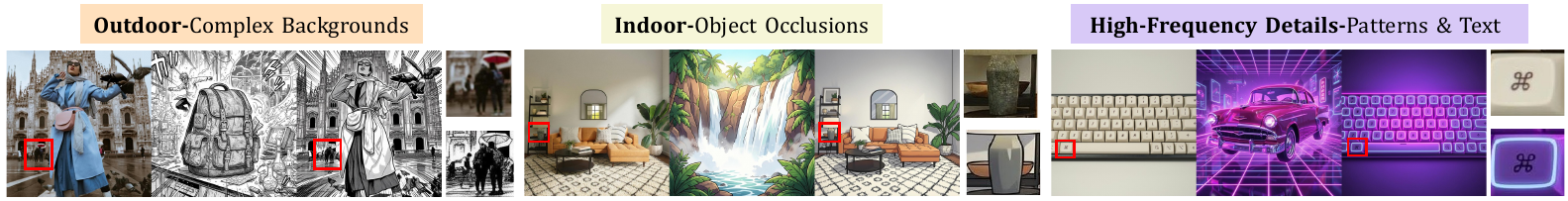}
    % \vspace{-2em}
    \caption{\textbf{Robustness of stylization in challenging scenarios.} Red bounding boxes and zoomed-in patches highlight precise structure preservation.}
    \label{fig:complex}
    % \vspace{-2em}
\end{figure*}

\noindent\textbf{Content-Style Trade-off Analysis.}
Fig.~\ref{fig:scatter} shows most methods exhibit a clear trade-off between content preservation and style fidelity, excelling in one dimension often comes at the expense of the other. In the style transfer comparison (left two), our method consistently occupies the upper-right region, achieving the best balance among all competitors. In the image editing comparison (right two), our method attains competitive performance close to Nano-Banana-Pro and GPT-Image-1.5, despite being a style transfer model. These results show that DeStyle-350K enables well-disentangled content-style representations.

\noindent\textbf{User Study.}
\label{subsec:user_study}
We conducted a user study to assess the quality of stylization results. Participants were shown outputs from all methods and asked to rank their top three favorites based on: (1) style preservation, how well the style of the reference image is reflected; (2) content preservation, the degree to which structural details of the content image are retained; and (3) aesthetic appeal, overall visual quality. To reduce bias, image order was randomized and zooming was enabled. We collected 1,620 votes from 30 participants. As shown in Table~\ref{tab:us_1} and Table~\ref{tab:us_2}, we report both Rank-1 proportions and Top-3 selection rates. Results show a clear preference for our method: it outperforms existing style transfer methods and achieves performance close to the SOTA closed-source image editing model.

\noindent\textbf{Further Analysis.} \textit{(1) Stylization in Complex Scenarios.} As illustrated in Fig.~\ref{fig:complex}, our model demonstrates exceptional robustness in various challenging scenarios. In Outdoor-Complex Backgrounds, it anchors sharp architectural lines and individual silhouettes despite dense crowds. For Indoor-Object Occlusions, the model successfully maintains the original spatial occlusion relationships, ensuring distinct contours for partially hidden entities. Finally, in High-Frequency Details-Patterns \& Text, it preserves strict regularity and stroke clarity without geometric warping. These results further demonstrate the value of our DeStyle-350K dataset in ensuring structural integrity during stylization. \textit{(2) Effectiveness of DeStyle-350K.} 
As shown in Table~\ref{tab:quantitative_results_backbone}, fine-tuning three diverse backbones on DeStyle-350K yields consistent improvements across all metrics. Most notably, Qwen-L drops dramatically (e.g., 0.4380$\to$0.0025 on Flux.2-Klein-4B), indicating that models learn to transfer style without leaking semantic content. The consistent gains across architectures of different scales and families confirm that DeStyle-350K provides universally effective and backbone-agnostic supervision.

\vspace{10em}
\section{Conclusion}
\label{sec:conclusion}
This paper introduces a scalable paradigm for supervised style transfer built on destylization. By reducing stylistic elements from artistic images to recover their natural counterparts, we enable unaltered artworks to serve as authentic ground-truth targets, effectively addressing the long-standing absence of reliable supervision.  The resulting DeStyle-350K dataset, comprising 350K triplets across over 500 style categories, provides authentic and scalable supervision for training. Extensive evaluations on BCS-Bench across multiple architectures demonstrate that models trained on DeStyle-350K achieve superior stylization quality while maintaining structural integrity. We hope that the destylization paradigm and DeStyle-350K will serve as a solid foundation for future research in style transfer and image editing.

\newpage

\bibliographystyle{splncs04}
\bibliography{main}

\end{document}